\documentclass[11pt]{article}

\usepackage{arxiv}

\usepackage{amsmath,amsfonts,latexsym,fullpage,graphicx,vmargin, mathrsfs}
\usepackage{pstricks}
\usepackage{comment}
\usepackage[inline]{enumitem}
\usepackage{cleveref}

\usepackage[ruled]{algorithm2e}
\usepackage{algpseudocode}

\RequirePackage[OT1]{fontenc}

\usepackage{amsmath}
\usepackage{amsfonts}
\usepackage{caption}
\usepackage{subcaption}
\usepackage{multirow}
\usepackage{tikz}
\usetikzlibrary{graphs}
\usetikzlibrary{cd}
\usepackage{bm}
\usepackage{url}
\usepackage{placeins}

\RequirePackage{natbib}

\newtheorem{teo}{Theorem}
\newtheorem{prop}{Proposition}
\newtheorem{defi}{Definition}
\newtheorem{rmk}{Remark}

\newtheorem{cor}{Corollary}

\DeclareMathOperator*{\pers}{Pers}
\DeclareMathOperator*{\relu}{ReLu}

\usepackage{scalerel}

\newcommand{\Ss}{\mathbb{S}}
\newcommand{\R}{\mathbb{R}}

\newcommand{\N}{\mathbb{N}}

\newcommand{\virgolette}[1]{``#1''}

\firstpageno{1}

\begin{document}

\title{Persistence Spheres: Bi-Continuous Representations of Persistence Diagrams.}

\author{Matteo Pegoraro\thanks{Department of Mathematics, KTH}}

%
\maketitle

\begin{abstract}
We introduce persistence spheres, a novel functional representation of persistence diagrams. Unlike existing embeddings—such as persistence images, landscapes, or kernel methods—persistence spheres provide a bi-continuous mapping: they are Lipschitz continuous with respect to the 1-Wasserstein distance and admit a continuous inverse on their image. This ensures, in a theoretically optimal way, both stability and geometric fidelity, making persistence spheres the representation that most closely mirrors the Wasserstein geometry of PDs in linear space. 
We derive explicit formulas for persistence spheres, showing that they can be computed efficiently and parallelized with minimal overhead. Empirically, we evaluate them on diverse regression and classification tasks involving functional data, time series, graphs, meshes, and point clouds. Across these benchmarks, persistence spheres consistently deliver state-of-the-art or competitive performance compared to persistence images, persistence landscapes, and the sliced Wasserstein kernel.
\end{abstract}

\begin{keywords}
Topological Data Analysis, Persistence Diagrams, Lift Zonoid, Vectorization, Topological Machine Learning
\end{keywords}

\section{Introduction}

Topological Data Analysis (TDA) is an emerging field that leverages concepts from algebraic topology to study the shape of data, offering coordinate-free and noise-robust methods for extracting meaningful patterns. At the core of TDA lies persistent homology, a framework that captures multi-scale topological features of a dataset. By recording the scales at which features such as connected components, loops, and voids appear (birth) and disappear (death), persistent homology produces compact descriptors of data shape. These descriptors are commonly represented as persistence diagrams (PDs) or barcodes, which provide stable and interpretable summaries amenable to qualitative exploration and (limited) quantitative analysis \citep{edelsbrunner2010computational, oudot2015persistence}.

\paragraph{Data Analysis with Persistence Diagrams.}
To integrate topological information into data analysis pipelines, PDs are often compared using Wasserstein distances defined through partial optimal transport (POT) \citep{divol2021understanding}. These distances play a crucial role in ensuring robustness to perturbations, but they also impose a highly non-linear geometry on the space of PDs. This non-linearity significantly limits the range of statistical tools that can be directly applied to PDs. For instance, even basic operations such as computing averages are non-trivial: they are usually formulated in terms of Wasserstein barycenters \citep{mileyko2011probability}, which are computationally intensive to approximate and may fail to yield unique solutions.

\paragraph{Topological Machine Learning: Vectorizations and Kernel Methods.}
To overcome these limitations, numerous vectorization methods have been developed to embed PDs into linear spaces, enabling the use of classical statistical and machine learning techniques. Such embeddings underpin the field of \emph{topological machine learning} \citep{papamarkou2024position}, where topological features and topological loss functions have proven effective in both predictive and representation learning tasks \citep{moor2020topological, waylandmapping}. For comprehensive surveys we refer to \cite{pun2022persistent, ali2023survey, papamarkou2024position}; here we only recall the main approaches.

Broadly, these methods fall into two main categories. The first consists of explicit embeddings of PDs into linear spaces, while the second comprises kernel methods \citep{reininghaus2015stable, kusano2018kernel, carriere2017sliced}, which employ the \emph{kernel trick} to define feature maps implicitly. Within the class of explicit embeddings, one can further distinguish between approaches based on \emph{descriptive statistics} \citep{asaad2022persistent}, \emph{algebraic representations} exploiting polynomial rings or tropical coordinates \citep{kalivsnik2019tropical, monod2019tropical, di2015comparing}, \emph{functional representations}, which associate to each diagram a scalar field over a chosen domain \citep{bubenik2015statistical, adams2017persistence, biscio2019accumulated, gotovac2025topological} and other approaches \citep{mitra2024geometric}.

\paragraph{Main Contributions.}
In this work, we build on the framework of \cite{gotovac2025topological} (see \Cref{rmk:gotovac}) and propose a new functional representation of PDs, mapping a diagram $D$ to a function $\varphi:\mathbb{S}^2 \to \mathbb{R}$. We show that this mapping is Lipschitz continuous with respect to the 1-Wasserstein distance between diagrams, and that its inverse—on its domain of definition—is also continuous. Continuity of the forward map guarantees stability, since similar diagrams yield similar functions, while continuity of the inverse ensures that functional similarity always corresponds to similarity between diagrams.
Crucially, this bi-continuity establishes the strongest possible geometric correspondence between the Wasserstein space of PDs and their functional representation, given that a bi-Lipschitz embedding is known to be impossible \citep{carriere2019metric}. 
To the best of our knowledge, no other vectorization of PDs combines these properties.
Stronger guarantees can only be obtained by restricting attention to diagrams with at most $n$ points, for some fixed $n$, as in \cite{mitra2024geometric}.

\section{Preliminaries} 

\subsection{Convex Sets and Support Functions}

We briefly review the notation and concepts from convex analysis and geometry that will be used throughout. Standard references include \cite{rockafellar1997convex, salinetti_convex}.

\begin{defi}
    Given two convex sets $A,B\subset \R^2$, their Minkowski sum and their multiplication with a non-negative scalar $\lambda \geq 0$,  are defined as:
    \[
    A\oplus B = \{a+b \mid a\in A,b\in B\}, \\
    \lambda A = \{\lambda a \mid a\in A\}.
    \]
\end{defi}

\begin{defi}\label{defi:support}
    Given a compact convex set $A\subset \R^2$, its support function is defined as:
    \begin{align*}
    h_A:\R^2&\rightarrow \R \\
    x &\mapsto \max_{a\in A} \langle x,a\rangle.
    \end{align*}
\end{defi}

One can check that 1) any support function is completely determined by its restriction on $\Ss^2$; 2) the operator $A\mapsto h_A$ is linear: 
$\lambda_1 A \oplus \lambda_2 B \mapsto \lambda_1 h_A + \lambda_2 h_B$.

To compare different convex sets we will use the Hausdorff distance.

\begin{defi}
    Given two compact subsets $A,B\subset Z$, with $(Z,d_Z)$ being a metric space, their Hausdorff distance is defined as:
    \[
    d_H(A,B)= \max \lbrace \max_{a\in A}d_Z(a,B),\max_{b\in B}d_Z(b,A)\rbrace
    \]
\end{defi}

Now we can state the following classical result.

\begin{prop}\label{prop:convex}
    Given two compact convex sets $A,B\subset \R^2$, the following holds:
\[
\max_{v\in \Ss^2}\parallel h_A(v)-h_B(v) \parallel_2 = d_H(A,B).
\]
In particular, the operator $A\mapsto h_A$ is injective.
\end{prop}

\subsection{Integrable Measures on $\R^2$}

For any Borel measure $\mu$ on $\R^2$, and any $f:\R^2\rightarrow \R$ $\mu$-measurable, we set:
    \[
 \langle\mu,f\rangle := \int_{\R^2} f(p) d\mu(p).
    \]
Moreover, for any $r\geq 0$, we set $B_r = \{p\in \R^2 \mid \parallel p \parallel_2 \leq r\}$, and $B_r^c = \R^2\setminus B_r$.

In the following we will use \emph{integrable} measures and \emph{uniformly} integrable sequences of measures. See \cite{hendrych2022note} for more details on such topics.

 \begin{defi}
    A positive finite Borel measure on $\R^2$, $\mu$, is called integrable if:
    \[
    \langle\mu,\parallel \cdot\parallel_2\rangle = \int_{\R^2} \parallel p\parallel_2 d\mu(p)<\infty .
    \]
    Similarly, a sequence of integrable measures $\{\mu_n\}_{n\in \N}$ is uniformly integrable if:
    \[
    \lim_{r\rightarrow \infty} \sup_{n\in \N} \int_{B_r^c} \parallel p\parallel_2 d\mu_n(p) =0.
    \]
\end{defi}

To compare measures, we need weak and vague convergence of measures, which are standard notions in measure theory. See, for instance, \cite{kallenberg1997foundations}.

\begin{defi}
  A sequence of integrable measures $\{\mu_n\}_{n\in \N}$ converges weakly to $\mu$ if $\langle\mu_n,f\rangle \rightarrow\langle\mu,f\rangle$
  for every $f:\R^2\rightarrow \R$ continuous and bounded. Instead, if $\langle\mu_n,f\rangle \rightarrow\langle\mu,f\rangle$ for  every $f:\R^2\rightarrow \R$ continuous and compactly supported, we say that $\{\mu_n\}_{n\in \N}$ converges vaguely to $\mu$. 
\end{defi}
 
 We write $\mu_n \xrightarrow{w} \mu$ for weak convergence and $\mu_n \xrightarrow{v} \mu$ for vague convergence.  

\subsection{Persistence Diagrams}

We adopt a measure-theoretic perspective to define PDs. 
First, we introduce the following notation:
    \[
    \R^2_{x<y}:=\{(x,y)\in \R^2\mid x<y\},\text{       } \\ \Delta := \{(x,y)\in \R^2\mid x=y\}.
    \]

\begin{defi}
    A PD is a positive finite measure $\mu_D=\sum_{p\in D}c_p \delta_p$, with $\delta_p$ being the Dirác delta centered in $p\in\R^2$, $D\subset\R^2_{x<y}$ being a finite set, and $c_p\in \N$. We refer to the set $D$ as the support of the diagram.
\end{defi}

Following \cite{divol2021understanding} we give the following definition.

\begin{defi}
    For any measure $\mu$ and for any subset $Z\subset \R^2$, we define:
    \[
    \textstyle\pers_Z(\mu) = \displaystyle\frac{1}{2}\int_{Z} (y-x) d\mu((x,y)).
    \]
    When $Z=\R^2$, we simply write $\pers(\mu)$.
\end{defi}

As proven in \cite{skraba2020wasserstein}, in the context of stability for linear operators defined on spaces of measures, we are forced to work with the $1$-Wasserstein metric.
To introduce such a metric with a notation convenient for the proofs that follow, we define the following terms. 

\begin{defi}
    Consider two diagrams $\mu_D$ and $\mu_{D'}$.
A partial matching between $\mu_D = \sum_{p\in D} a_p \delta_p$ and $\mu_{D'}=\sum_{p\in D'}b_p \delta_p$ is a triplet $(D_\gamma,D'_\gamma,\gamma: D_\gamma\rightarrow D'_\gamma)$ such that:
\begin{itemize}
    \item $D_\gamma\subset D$ and $D_\gamma'\subset D'$;
    \item $\gamma: D_\gamma\rightarrow D'_\gamma$ is a bijection.
\end{itemize}
\end{defi}

We may indicate a partial matching just with $\gamma$, for the sake of brevity.

Given a partial matching $\gamma$ between $\mu_D = \sum_{p\in D} a_p \delta_p$ and $\mu_{D'}=\sum_{p\in D'}b_p \delta_p$, for every $p\in D_\gamma$, we set $\gamma_p := \min \{a_p, b_{\gamma(p)}\}$. Similarly, for every $q\in D'_\gamma$, we set $\gamma_q := \min \{b_q, a_{\gamma^{-1}(q)}\}$. The cost of $\gamma$ can then be defined as follows:
\begin{equation}
\begin{aligned}
    c(\gamma):= & \sum_{p\in D_\gamma} \gamma_p\parallel p-\gamma(p) \parallel_\infty + \sum_{p\in D_\gamma} (a_p-\gamma_p)\parallel p-\Delta \parallel_\infty + \sum_{q\in D'_\gamma} (b_q-\gamma_q)\parallel q-\Delta \parallel_\infty +\\
    &\sum_{p\in D\setminus D_\gamma} a_p\parallel p-\Delta \parallel_\infty + \sum_{q\in D'\setminus D'_\gamma} b_q\parallel q-\Delta \parallel_\infty.    
\end{aligned}
\end{equation}

\begin{defi}
The $1$-Wasserstein distance between PDs is defined as:
    \[
    W_1(\mu_D,\mu_{D'}) = \inf \{c(\gamma)\mid \gamma \text{ partial matching between }\mu_D\text{ and }\mu_{D'}\}.
    \]
\end{defi}

\begin{rmk}
The definition of the $1$-Wasserstein distance adopted here is equivalent to other formulations in the literature. 
In particular, the bijection 
\(\gamma : D_\gamma \to D'_\gamma\) 
can be interpreted as a transport map between the measures 
$\sum_{p \in D_\gamma} \gamma_p \, \delta_p $
and  
$\sum_{q \in D'_\gamma} \gamma_q \, \delta_q$, 
while the associated cost \(c(\gamma)\) corresponds to the transportation cost, including the cost of sending the remaining mass of both diagrams to the diagonal \(\Delta\) (see \cite{divol2021understanding}). 
\end{rmk}

We recall the following key result from  \cite{divol2021understanding}.

\begin{teo}\label{prop:wass_conv}
      \[
      W_1(\mu_D,\mu_{D_n})\rightarrow 0 \text{ if, and only if, } \mu_{D_n} \xrightarrow{v} \mu_D  \text{ and } \pers(\mu_{D_n})\rightarrow \pers(\mu_D).
      \]
\end{teo}

\subsection{Lift Zonoids of Discrete Measures}

We now introduce the final components needed to define our topological summaries. 
Throughout, we adopt the following notation: for a point \(p=(x,y)\in \R^2\), we set 
\((1,p):=(1,x,y)\in \R^3\).

As a preliminary step, we recall the construction of the lift zonoid associated with an integrable measure, as presented in \cite{koshevoy1998lift, hendrych2022note}. For simplicity and coherence with our setting, we restrict attention to discrete measures.

\begin{defi}
    Given a discrete measure $\mu = \sum_{i=1}^n c_i\delta_{p_i}$, $p_i\in \R^2$ and $c_i>0$, the lift zonoid of $\mu$ is the following convex set (zonotope):
    \[
       Z_\mu = \bigoplus_{i=1}^n c_i [0,(1,p_i)]\subset \R^{3},    
    \]
    with $[0,(1,p_i)]$ being the segment joining the origin $0\in \R^3$ and the point $(1,p_i)$.
\end{defi}

Note that the lift zonoid construction is linear: $\lambda_1 \mu_1 + \lambda_2 \mu_2 \mapsto \lambda_1 Z_{\mu_1} \oplus \lambda_2 Z_{\mu_2}$

\cite{koshevoy1998lift, hendrych2022note} prove the following result.

\begin{prop}
    Given an integrable measure $\mu$ and a sequence of integrable measures $\{\mu_n\}_{n\in \N}$, the following hold: 
      \[
      d_H(Z_{\mu},Z_{\mu_n})\rightarrow 0 \text{ if, and only if, } \mu_{n} \xrightarrow{w} \mu  \text{ and } \{\mu_{n}\}_{n\in\N} \text{ is uniformly integrable}.
      \]
\end{prop}

\section{Persistence Spheres}\label{sec:pers_spheres}

Now we finally introduce persistence spheres as the support functions (see \Cref{defi:support}) of lift zonoids of (weighted) PDs, restricted to $\Ss^2$.

As for other functional representations of PDs, see \cite{adams2017persistence}, we need to re-weight diagrams with a function $\omega:\R^2\rightarrow (0,1]$ so that the weight assigned to points goes to zero as we approach $\Delta$.
Given a diagram $\mu_D=\sum_{p\in D}c_p \delta_p$ and a function $\omega:\R^2\rightarrow (0,1]$, we set $\mu_D^\omega:= \sum_{p\in D} \omega(p) c_p \delta_p$.

\begin{defi}
    Given a PD $\mu_D$ and a function $\omega:\R^2\rightarrow (0,1]$, the persistence sphere (PS) of $\mu_D$ with weighting $\omega$ is defined as $\varphi_{\mu_D}^\omega:=(h_{Z_{\mu^\omega_D}})_{\mid \Ss^2}$.
\end{defi}

For any function $\omega:\R^2\rightarrow (0,1]$ we set $\Gamma_\omega(p):=\omega(p) (1,p)$.
We want to control the decay of $\omega$ as points approach the diagonal. We do so with the following technical conditions.

\begin{defi}\label{def:stable_w}
    A function $\omega:\R^2\rightarrow (0,1]$ is called a stable lift weighting if: 
\begin{itemize}
    \item  $\Gamma_\omega$ is $C$-Lipschitz for some $C>0$;
    \item the following inequality is satisfied for every $p=(x,y)\in \R^2_{x<y}$ and some fixed $C'>0$:
        \[
    \parallel\Gamma_\omega(p)\parallel_2 \leq C'  \left(\frac{y-x}{2}\right) = C' \parallel p-\Delta \parallel_\infty.
    \]
\end{itemize}
\end{defi}

In \Cref{def:stable_w}, we used the term ``lift weighting'' to emphasize its role in the context of lift zonoids. Since no ambiguity arises in this work, we will omit the qualifier ``lift'' from now on for brevity.

\begin{defi}\label{defi:effective}
    A function $\omega:\R^2\rightarrow (0,1]$ is called an effective (lift) weighting if for any sequence of diagrams $\{\mu_{D_n}\}_{n\in\N}$:  
\[
\lim_{r\rightarrow \infty} \sup_n \int_{B_r^c} \omega(p)\,\|p\|_2 \, d\mu_n(p)=0
\quad \implies \quad
\lim_{r\rightarrow \infty} \sup_n \textstyle\pers_{B_r^c}(\mu_{D_n})=0.
\]
\end{defi}

\Cref{defi:effective} controls the behavior of $\Gamma_\omega$ at infinity. To see that, note that, for every $\varepsilon>0$, there is $R$ such that for every $r>R$ we have: $\frac{\parallel p\parallel_2 }{\parallel (1,p)\parallel_2 } \geq 1-\varepsilon$. For any such $r$:
\[
(1-\varepsilon)  \sup_n \int_{B_r^c} \omega(p) \parallel (1,p)\parallel_2 d\mu_{D_n}(p)  \leq \sup_n \int_{B_r^c} \omega(p) \parallel p\parallel_2 d\mu_{D_n}(p).
\]
Which means that, in the context of the definition, we have:
\begin{equation}\label{eq:effective}
\lim_{r\rightarrow \infty} \sup_n \int_{B_r^c} \parallel  \Gamma_\omega(p) \parallel_2 d\mu_{D_n}(p)\rightarrow 0.    
\end{equation}

We now provide examples of stable and effective weightings.

\begin{prop}
Set $\lambda(p) := \frac{y-x}{2\parallel (1,p) \parallel_2}$. The following are stable weightings:
\[
\widetilde{\omega}(p)=\lambda(p)^\alpha, 
\qquad
\omega_K(p)=\frac{2}{\pi}\arctan \left(\frac{\lambda(p)^\alpha}{K^\alpha} \right), 
\]
 for any $K>0$ and $\alpha\geq1$. They are also effective weightings for $\alpha=1$. 
\end{prop}

We conclude this section highlighting that, by linearity, the PS of a PD $\mu_D=\sum_{p\in D} c_p \delta_p$, with weighting function $\omega$, can be explicitly written as:
    \begin{equation}\label{eq:PS}
    \varphi_{\mu_D}^\omega(v)=h_{Z_{\mu^\omega_D}}(v) = \sum_{p \in D} \omega(p) c_p  \relu(\langle v,(1,p)\rangle ),    \qquad  \text{ with }  \relu(x):=\max\{0,x\}.
    \end{equation}

\subsection{Continuity Theorems}

We now state our main results, which contain the continuity properties anticipated in the introduction. 
First we state and prove them in terms of lift zonoids and Hausdorff distances, which simplifies the proofs, and then, using \Cref{prop:convex}, we derive the bi-continuity of PSs. 

\begin{teo}\label{teo:stable}
    Let $\mu_D,\mu_D'$ be PDs and let $\omega:\R^2\rightarrow \R$  be a stable weighting.
    We have:
    \[
    d_H(Z_{\mu^{\omega}_D},Z_{\mu^{\omega}_{D'}}) \leq \max\{C,C'\}\cdot W_1(\mu_D,\mu_{D'}),
    \]
    with $C,C'>0$ being the stability constants of $\omega$ (see \Cref{def:stable_w}).
\end{teo}

\begin{teo}\label{teo:inverse}
    Let $\{\mu_{D_n}\}_{n\in \N}$ be a sequence of PDs such that $d_H(Z_{\mu^{\omega}_{D_n}},Z_{\mu^{\omega}_{D}}) \rightarrow 0$, with $\omega:\R^2\rightarrow \R$  being an effective weighting. Then, $W_1(\mu_{D_n},\mu_D)\rightarrow 0$.
\end{teo}

Summarizing the statements of \Cref{teo:stable} and \Cref{teo:inverse}, and writing them replacing lift zonoids with persistence spheres, we obtain the following result. 

\begin{cor}
    Within the setting of the previous results, we have:
\begin{itemize}
    \item for every $p \in [1,\infty]$ there exist $C_p>0$ such that, for every pair of diagrams $\mu_D,\mu_{D'}$, we have $\parallel \varphi_{\mu_D}^\omega- \varphi_{\mu_{D'}}^\omega \parallel_p \leq C_p W_1(\mu_D,\mu_{D'})$;
    \item if $\parallel \varphi_{\mu_D}^\omega- \varphi_{\mu_{D_n}}^\omega \parallel_\infty \rightarrow 0$, then $W_1(\mu_{D_n},\mu_D)\rightarrow 0$.
\end{itemize}

\end{cor}

\begin{rmk}\label{rmk:gotovac}
\cite{gotovac2025topological} define their functional representation as $\varphi_{\mu_D}^\omega$, with $\omega(p) = y - x$. This weighting is not stable, since for any $C > 0$,
\begin{equation}\label{eq:gotovac}
   \| \Gamma_\omega(p) \| > C \|p - \Delta \|_\infty,
\end{equation}
whenever $\|p\|_2$ is sufficiently large. In particular, the map $\mu_D \mapsto \varphi_{\mu_D}^\omega$ fails to be stable. For instance, let $D_n = \{p_n\} = \{(n^2, n^2 + \tfrac{1}{n})\}$. Then
$Z_{D_n^\omega} = \tfrac{1}{n}[0, (1, p_n)].$
If $D = \emptyset$, we obtain
\[
W_1(D_n, D) = \tfrac{1}{n} \xrightarrow{n \to \infty} 0,
\qquad 
d_H(Z_D, Z_{D_n}) \geq \tfrac{\sqrt{2}}{n}\, n^2 \xrightarrow{n \to \infty} \infty.
\]

On the other hand, by \Cref{eq:gotovac} one can verify that $\omega(p) = y - x$ is effective. Consequently, \Cref{teo:inverse} holds for the functional representation in \cite{gotovac2025topological}, although this fact is not established in that work.
\end{rmk}

\section{Experiments}
\label{sec:experiments}

\begin{figure}
\centering
	\begin{subfigure}[c]{0.3\textwidth}
    	\includegraphics[width = \textwidth]{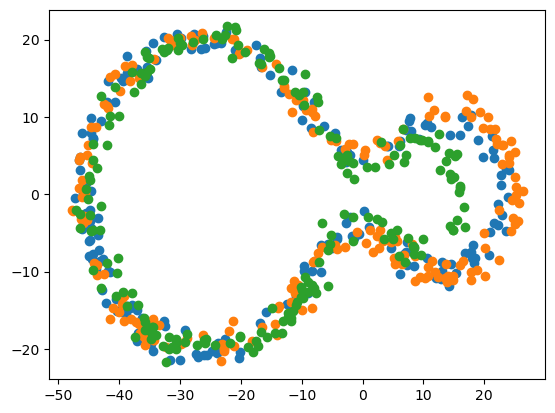}
		\captionsetup{singlelinecheck=off, margin={0.05cm, 0.05cm}}
    	\caption{Plot of three point clouds in the \virgolette{Eyeglasses} case study.}
    	\label{}
    \end{subfigure}
    	\begin{subfigure}[c]{0.3\textwidth}
    	\centering
    	\includegraphics[width = \textwidth]{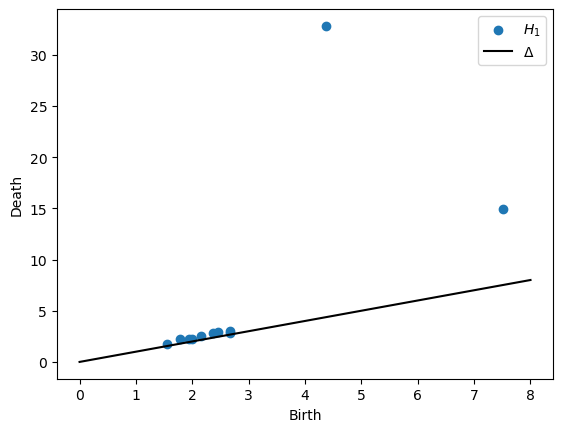}
		\captionsetup{singlelinecheck=off, margin={0.05cm, 0.05cm}}
    	\caption{A PD from the \virgolette{Eyeglasses} case study.}
    	\label{fig:PD}
    \end{subfigure}
    	\begin{subfigure}[c]{0.3\textwidth}
    	\centering
    	\includegraphics[width = \textwidth]{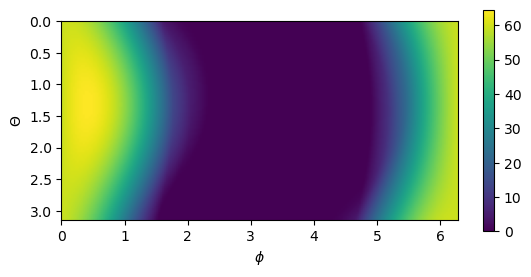}
		\captionsetup{singlelinecheck=off, margin={0.05cm, 0.05cm}}
    	\caption{The PS of the PD in \Cref{fig:PD}, from the \virgolette{Eyeglasses} case study, in polar coordinates.}
    	\label{}
    \end{subfigure}

	\begin{subfigure}[c]{0.3\textwidth}
		\centering
		\includegraphics[width = \textwidth]{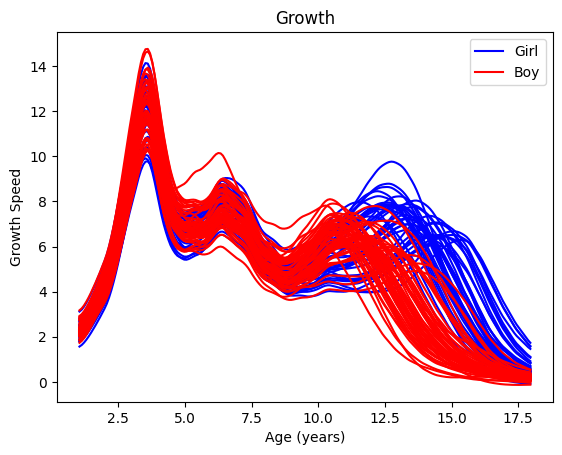}
		\captionsetup{singlelinecheck=off, margin={0.05cm, 0.05cm}}
		\caption{The (derivatives of the) \virgolette{Growth} dataset, with the two classes highlighted.}
		\label{fig:growth}
	\end{subfigure}
	\begin{subfigure}[c]{0.3\textwidth}
		\centering
		\includegraphics[width = \textwidth]{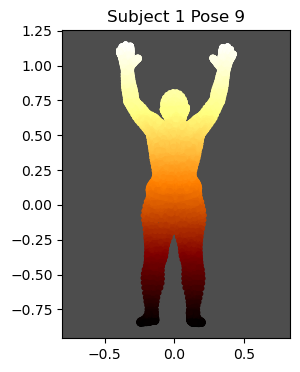}
		\captionsetup{singlelinecheck=off, margin={0.05cm, 0.05cm}}
		\caption{A sample from the \virgolette{Human Poses} dataset.}
		\label{fig:human}
	\end{subfigure}	\begin{subfigure}[c]{0.3\textwidth}
		\centering
		\includegraphics[width = \textwidth]{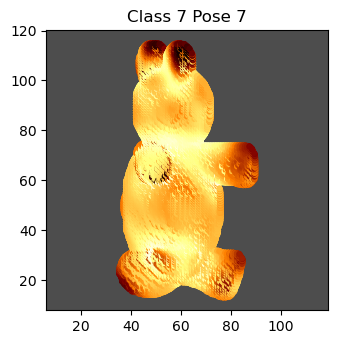}
		\captionsetup{singlelinecheck=off, margin={0.05cm, 0.05cm}}
		\caption{A sample from the \virgolette{Mc Gill 3D Shapes} dataset.}
		\label{fig:mcgill}
	\end{subfigure}
    \caption{Data, PDs, and PSs from some of the experiments in \Cref{sec:experiments}.}
    \label{fig:experiments}
\end{figure}

We evaluated PSs on a range of regression and classification case studies, comparing their performance with persistence images (PIs), persistence landscapes (PLs), and the sliced Wasserstein kernel (SWK). For PSs, PIs, and PLs, we employed both support vector machines (SVM) with radial-basis kernels (denoted \virgolette{-SVM} in \Cref{table:results}) and penalized linear/logistic regressions (denoted \virgolette{-Reg} in \Cref{table:results}). 
Performance was assessed using $R^2$ for regression tasks and  accuracy for classification tasks, averaged over $10$ independent runs. See also \Cref{fig:experiments}.

\subsection{Datasets}

We now briefly describe the case studies under consideration, organizing them according to the source of the datasets. Our selection draws from a diverse range of origins: some datasets were generated by us, others stem from benchmark case studies in functional data analysis (FDA) and shape analysis, and additional ones were taken from publications where PDs were publicly available—allowing us to bypass a full pipeline implementation. In making these choices, we also prioritized diversity in the underlying data types, including functions, time series, 3D meshes, graphs, and point clouds.

\paragraph{\virgolette{Eyeglasses} Case Study}

The \virgolette{Eyeglasses} dataset is a regression case study we designed using the \emph{eyeglasses} generative model from the \url{scikit-tda} python package \citep{scikittda2019}. This model takes two radii as parameters, and a noise variable which was kept equal to $1$. The first radius was always set equal to $20$, while the second was sampled according to a normal distribution with mean $10$ and standard variation $2.5$. We sampled $2000$ point clouds and used a $30\%-70\%$ split between training and test data; threefold cross-validation was used to select hyper-parameters, and $1$-dimensional PDs where obtained from the Vietoris-Rips filtration.

\paragraph{Functional datasets from the \url{scikit-fda} Package}

For the following functional datasets, we used zero-dimensional persistent homology derived from the sublevel set filtration. Data were split into training and test sets in a $70\%–30\%$ ratio, and hyperparameters were selected via threefold cross-validation. All datasets are freely available in the \texttt{scikit-fda} Python package \citep{scikitfda}.

The \virgolette{Tecator} dataset (\url{https://lib.stat.cmu.edu/datasets/tecator}) consists of publicly available measurements collected using the \virgolette{Tecator Infratec Food and Feed Analyzer}. Building on the derivatives of these curves, we explore the same regression problem as in \cite{ferraty_fda}, trying to regress the fat content of the food samples.

The \virgolette{$\text{NO}_x$} dataset \citep{febrero2008outlier} 
contains hourly measurements of daily nitrogen oxides ($\text{NO}_x$) emissions in the Barcelona area. The data is labeled based on whether the emission curve was recorded on a weekday or a weekend, and our goal is thus to reconstruct this labeling through supervised classification.

    The \virgolette{Growth} dataset \citep{tuddenham1954physical}, also known as \virgolette{The Berkeley Growth Study}, contains height measurements of girls and boys, recorded yearly between ages 1 and 18. A common approach is to analyze the first derivative of the growth curves to distinguish growth dynamics between boys and girls \citep{vitelli2010functional}.

\paragraph{Datasets from \cite{bandiziol2024persistence}}

The classification case studies involving the datasets \virgolette{DYN SYS}, \virgolette{ENZYMES JACC}, \virgolette{POWER}, and \virgolette{SHREC14} were taken from \cite{bandiziol2024persistence}. As in the previous setting, we used a $70\%–30\%$ train–test split, with hyperparameters selected via threefold cross-validation. For these datasets, we could directly rely on the PDs associated with the classification tasks, which are publicly available at \url{https://github.com/cinziabandiziol/persistence_kernels}.

In selecting the problems, we prioritized classification tasks with balanced classes and diversity in data type, including point clouds, graphs, time series, and 3D meshes. We now summarize the considered datasets; further details can be found in \cite{bandiziol2024persistence}.

The dataset \virgolette{DYN SYS}, first introduced in \cite{adams2017persistence} and referred to as \virgolette{Orbit Recognition} in \cite{bandiziol2024persistence}, consists of point clouds generated by a one-parameter discrete dynamical system, with the parameter ranging in $\{2.5, 3.5, 4, 4.1, 4.3\}$. The classification task, considered in \cite{adams2017persistence, bandiziol2024persistence} as well as in our work, is to predict the parameter value from the associated point cloud, a problem also studied in \cite{carriere2017sliced}. For each parameter value, $50$ independent point clouds were generated, each containing $1000$ points with starting positions chosen uniformly at random, yielding a dataset of $250$ elements. The PDs computed by \cite{bandiziol2024persistence} contain only one-dimensional features.

The dataset \virgolette{ENZYMES JACC} addresses a graph classification problem. Graphs represent protein tertiary structures obtained from the BRENDA enzyme database (\url{https://www.brenda-enzymes.org/}), and the task is to classify each of the $600$ graphs into one of six enzyme classes. Edges were weighted by their Jaccard index, and PDs were computed from the resulting sublevel set filtration, combining both zero- and one-dimensional features.

The dataset \virgolette{POWER}, from the UCR Time Series Classification Archive (\url{https://www.cs.ucr.edu/~eamonn/time_series_data_2018/}), consists of 1096 time series. The pipeline in this case applied the sliding window embedding \citep{ravishanker2021introduction}, followed by the extraction of zero-, one-, and two-dimensional features, which were then merged into a single diagram for each time series.

Finally, the dataset \virgolette{SHREC14} \citep{Pickup2014} is a benchmark for non-rigid 3D shape classification. It contains meshes of human models across $20$ poses and $15$ body types (e.g., man, woman, child), resulting in $300$ total meshes. In \cite{bandiziol2024persistence}, the Heat Kernel Signature (HKS) \citep{sun2009concise, bronstein2010scale} was used to extract one-dimensional PDs from the corresponding sublevel set filtrations.

\paragraph{Datasets \virgolette{Human Poses} and \virgolette{Mc Gill 3D Shapes}}

The remaining datasets, \virgolette{Human Poses} and \virgolette{McGill 3D Shapes}, were obtained from \url{https://github.com/ctralie/TDALabs/blob/master/3DShapes.ipynb}. The corresponding classification pipelines are documented in the referenced notebook: for the human pose task, a sublevel set filtration of the height function was used, while for the McGill shape classification task, a sublevel set filtration of the HKS was applied. We note that the \virgolette{McGill 3D Shapes} dataset used here is a subsample of the original version, which is no longer fully accessible online. In both case studies, the train–test split $(80\%–20\%)$ was imposed by the dataset limitation of having only 10 samples per class.

\subsection{Implementation Details}
\label{sec:implementation_details}

As already mentioned, we considered two types of pipelines based on the chosen vectorization method: SVM  and penalized linear/logistic regression. Both were implemented using the \texttt{scikit-learn} Python package \citep{scikit-learn}. For regression and classification with linear/logistic models, we employed $\ell_2$ regularization to mitigate overfitting.

\paragraph{Regression and SVM Parameters}
First we consider hyperparameters associated with the learning models (\virgolette{-SVM} vs. \virgolette{-Reg}).

\begin{itemize}
    \item \virgolette{-SVM}: all SVM pipelines required a regularization parameter $C$, chosen from $\{0.001, 0.01, 0.1, 1, 10, 100, 1000, 10000\}$. Except for SWK, where the kernel was precomputed, all other methods used the radial basis function kernel.
    \item \virgolette{-Reg}: for Ridge regression, the penalization parameter $\alpha$ was chosen from $\{0.000001,0.00001,0.0001,0.001,0.01,0.1,1,10,100\}$, while for penalized logistic regression the $C$ parameter was selected from $\{1, 10, 100, 1000, 10000\}$.
\end{itemize}

\paragraph{Linearization Methods Parameters}
Now we consider hyperparameters tied to each linearization method.

\begin{itemize}
    \item PS: we employed the weighting function $\omega^\alpha_K$ defined in the main text, with $\alpha=1$, and exploring $K \in \{0.00001,0.0001,0.001,0.01,0.1,0.2,0.3,0.4,0.5\}$. Since PS are functions on $\Ss^2$ expressed in spherical coordinates, we projected them onto a spline basis and applied functional principal component analysis (FPCA) on the sphere—retaining all components—to obtain an orthonormal representation suitable for penalized models in \texttt{scikit-learn}. FPCA was implemented using the Rayleigh quotient approach, which accommodates non-orthonormal bases. 
    We used a grid of size $200\times 100$ to evaluate PSs, and a spherical spline basis with $561$ elements for the FPCA.
    \item PI: using the \texttt{scikit-tda} \url{persim} module, we selected the $pixel\_size$ by enclosing all PDs in a rectangle (expressed in birth–persistence coordinates) and dividing its shortest side by $n’$, with $n’ \in \{100, 500\}$, rounding to the nearest power of 10 for numerical stability. We used the default bivariate Gaussian kernel, with $\sigma = pixel\_size/m$, where $m \in \{1, 10, 100\}$. The exponent $n$ in the $weight\_params$ dictionary (corresponding to persistence weighting) was varied in $\{1,2,4,8\}$.
    \item PL: for landscapes, we considered the first five landscapes evaluated on a fixed common grid across the dataset, and we concatenated them. No hyperparameters needed tuning.
    \item SWK: we used the implementation of the sliced Wasserstein kernel in the \texttt{gudhi} library \citep{gudhi}, fixing $M=100$ and exploring the scaling parameter $\sigma \in \{0.00001, 0.0001, 0.001, 0.01, 0.1, 1, 10\}$ for the Gram matrix.
\end{itemize}

\paragraph{Computational Aspects}
The computation of a PS scales linearly with both the number of points in the diagram and the size of the evaluation grid. As shown in \Cref{eq:PS}, it reduces to evaluating standard mathematical functions in one or two variables. Since these evaluations are independent across points, the process can be efficiently parallelized with $O(1)$ work per core. As a result, PSs are potentially cheaper than PIs, which require binning and integration, and PLs, whose fastest known algorithm has complexity $O(n \log n + nN)$ \citep{bubenik2017persistence}, where $n = \#D$ and $N$ is the number of nonzero landscapes. Approximating SWK incurs a similar computational cost of $O(n \log n)$ \citep{carriere2017sliced}. When evaluated on a grid, PSs have the same dimensionality as PIs on a comparable grid, since both are scalar fields on 2D manifolds.  

Working with scalar fields on the sphere poses challenges for standard norm-penalized statistical methods (e.g., Ridge regression), which typically assume either a uniform grid or an orthonormal basis projection. To address this, we represented each PS using a spline basis and applied functional FPCA to obtain an orthonormal representation. While this approach increased computational cost, runtimes remained manageable in our experiments. For substantially larger datasets in combination with penalized methods, tailored implementations could further improve efficiency.

Finally, as with PIs, modifying the weighting function $\omega$ requires recomputing the entire persistence sphere. This differs from SWK, where the Gram matrix remains fixed across different values of $\sigma$.

\begin{table}[t]
\resizebox{\linewidth}{!}{
\begin{tabular}{llllllll}
\hline
\multicolumn{1}{|l|}{}                      & \multicolumn{1}{l|}{PSph-SVM}               & \multicolumn{1}{l|}{PSph-Lin}               & \multicolumn{1}{l|}{PI-SVM}                 & \multicolumn{1}{l|}{PI-Lin}        & \multicolumn{1}{l|}{PL-SVM}                 & \multicolumn{1}{l|}{PL-Lin}                 & \multicolumn{1}{l|}{SWK}                   \\ \hline
\textbf{Regression}                         &                                             &                                             &                                             &                                    &                                             &                                             &                                            \\ \hline
\multicolumn{1}{|l|}{Regression Case Study} & \multicolumn{1}{l|}{0.967}                  & \multicolumn{1}{l|}{\textbf{0.976}}         & \multicolumn{1}{l|}{0.672}                  & \multicolumn{1}{l|}{0.769}         & \multicolumn{1}{l|}{0.964}                  & \multicolumn{1}{l|}{0.883}                  & \multicolumn{1}{l|}{0.960}                 \\ \hline
\multicolumn{1}{|l|}{Tecator}               & \multicolumn{1}{l|}{\textbf{0.962 (0.004)}} & \multicolumn{1}{l|}{0.960 (0.007)}          & \multicolumn{1}{l|}{0.928 (0.029)}          & \multicolumn{1}{l|}{0.922 (0.018)} & \multicolumn{1}{l|}{0.939 (0.016)}          & \multicolumn{1}{l|}{0.837 (0.048)}          & \multicolumn{1}{l|}{0.953 (0.010)}         \\ \hline
\textbf{Classification}                     &                                             &                                             &                                             &                                    &                                             &                                             &                                            \\ \hline
\multicolumn{1}{|l|}{Growth}                & \multicolumn{1}{l|}{0.871 (0.064)}          & \multicolumn{1}{l|}{\textbf{0.889 (0.045)}} & \multicolumn{1}{l|}{0.525 (0.095)}          & \multicolumn{1}{l|}{0.536 (0.086)} & \multicolumn{1}{l|}{0.782 (0.051)}          & \multicolumn{1}{l|}{0.767 (0.107)}          & \multicolumn{1}{l|}{0.768 (0.058)}         \\ \hline
\multicolumn{1}{|l|}{NOx}                   & \multicolumn{1}{l|}{\textbf{0.877 (0.034)}} & \multicolumn{1}{l|}{0.826 (0.035)}          & \multicolumn{1}{l|}{0.803 (0.073)}          & \multicolumn{1}{l|}{0.563 (0.102)} & \multicolumn{1}{l|}{0.794 (0.047)}          & \multicolumn{1}{l|}{0.751 (0.040)}          & \multicolumn{1}{l|}{0.840 (0.055)}         \\ \hline
\multicolumn{1}{|l|}{DYN\_SYS}              & \multicolumn{1}{l|}{0.809 (0.040)}          & \multicolumn{1}{l|}{0.797 (0.030)}          & \multicolumn{1}{l|}{0.709 (0.020)}          & \multicolumn{1}{l|}{0.819 (0.017)} & \multicolumn{1}{l|}{\textbf{0.849 (0.029)}} & \multicolumn{1}{l|}{0.823 (0.027)}          & \multicolumn{1}{l|}{0.828 (0.028)}         \\ \hline
\multicolumn{1}{|l|}{ENZYMES\_JACC}         & \multicolumn{1}{l|}{\textbf{0.288 (0.064)}} & \multicolumn{1}{l|}{0.278 (0.036)}          & \multicolumn{1}{l|}{-}                      & \multicolumn{1}{l|}{-}             & \multicolumn{1}{l|}{0.236 (0.026)}          & \multicolumn{1}{l|}{0.254 (0.028)}          & \multicolumn{1}{l|}{0.283 (0.055)}         \\ \hline
\multicolumn{1}{|l|}{POWER}                 & \multicolumn{1}{l|}{0.761 (0.017)}          & \multicolumn{1}{l|}{0.738 (0.021)}          & \multicolumn{1}{l|}{0.682 (0.026)}          & \multicolumn{1}{l|}{0.700 (0.020)} & \multicolumn{1}{l|}{0.746 (0.014)}          & \multicolumn{1}{l|}{0.719 (0.030)}          & \multicolumn{1}{l|}{\textbf{0.767 (0.15)}} \\ \hline
\multicolumn{1}{|l|}{SHREC14}               & \multicolumn{1}{l|}{0.885 (0.057)}          & \multicolumn{1}{l|}{0.926 (0.019)}          & \multicolumn{1}{l|}{0.905 (0.029)}          & \multicolumn{1}{l|}{0.897 (0.024)} & \multicolumn{1}{l|}{0.914 (0.021)}          & \multicolumn{1}{l|}{\textbf{0.931 (0.013)}} & \multicolumn{1}{l|}{0.886 (0.092)}         \\ \hline
\multicolumn{1}{|l|}{Human Poses}           & \multicolumn{1}{l|}{0.580 (0.081)}          & \multicolumn{1}{l|}{\textbf{0.600 (0.084)}} & \multicolumn{1}{l|}{0.407 (0.144)}          & \multicolumn{1}{l|}{0.503 (0.046)} & \multicolumn{1}{l|}{0.460 (0.118)}          & \multicolumn{1}{l|}{0.480 (0.105)}          & \multicolumn{1}{l|}{0.345 (0.082)}         \\ \hline
\multicolumn{1}{|l|}{McGill 3D Shapes}      & \multicolumn{1}{l|}{0.383 (0.094)}          & \multicolumn{1}{l|}{0.489 (0.173)}          & \multicolumn{1}{l|}{\textbf{0.678 (0.060)}} & \multicolumn{1}{l|}{0.045 (0.052)} & \multicolumn{1}{l|}{0.650 (0.093)}          & \multicolumn{1}{l|}{0.511 (0.150)}          & \multicolumn{1}{l|}{0.567 (0.13)}          \\ \hline
\end{tabular}
}
\caption{Results of the case studies: we report average $R^2$ for regression and average accuracy for classification, across $10$ independent repetitions. Between brackets we reported the standard deviation of the scores. Best performing pipelines are reported in bold.}
\label{table:results}
\end{table}

\subsection{Results}

As shown in \Cref{table:results}, persistence spheres consistently matched or surpassed established topological representations—persistence images, persistence landscapes, and the sliced Wasserstein kernel—across regression and classification tasks spanning functional data, time series, graphs, meshes, and point clouds. The main exception was the McGill 3D Shapes dataset, where PIs and PLs performed better, possibly due the high variability induced by the limited number of samples per class.

\section{Conclusion and Broader Impact}

We introduced persistence spheres, a novel functional representation of persistence diagrams that is both Lipschitz continuous and admits a continuous inverse on its image, yielding a bi-continuous correspondence with respect to the 1-Wasserstein geometry. This combination of stability and geometric fidelity sets persistence spheres apart from existing vectorization methods. By providing a representation that is scalable, robust, and geometrically faithful, persistence spheres enable a more principled integration of topological information into machine learning pipelines. Empirically, we find that persistence spheres are not only competitive with, but frequently outperform, widely used alternatives such as persistence images, persistence landscapes, and sliced Wasserstein kernels.

Several avenues for future work remain. Alternative weighting schemes may yield more expressive summaries. Tools from functional data analysis could support advanced statistical methodologies, such as confidence sets, hypothesis testing, and limit theorems for point processes. Reconstruction techniques for recovering PDs from scalar fields on the sphere are under development. Visualization strategies could enhance interpretability. Integration, via differentiable loss functions, with modern representation learning techniques may broaden applicability. Finally, extending the construction to signed measures could provide a natural vectorization for bi-parameter persistence.

In summary, persistence spheres advance the state of topological machine learning by offering a theoretically rigorous, practically effective, and extensible framework for embedding topological information into learning systems.

\section*{Reproducibility Statement}
\Cref{sec:implementation_details} provides the main details required to reproduce our results. All datasets used are publicly available, and the explicit formulation of our method in \Cref{sec:pers_spheres} ensures reproducibility. 

\section*{The Use of Large Language Models}
Large Language Models were occasionally employed to refine and polish the writing.

\section*{Acknowledgments}
We thank C. A. N. Biscio, N. Chenavier, and N. R. Franco for the very helpful discussions.

\section*{Code}
The code employed to generate the results presented in this work has not yet been publicly released. We intend to make it available in the near future; in the meantime, it can be obtained upon request



\bibliography{iclr2026_conference}

\appendix

\section{Additional Implementation Details}
We conclude with a note on our current cross-validation pipeline for PSs. The code is designed to flexibly explore a wide parameter space (e.g., varying $\alpha$ rather than fixing $\alpha=1$), but this comes at the cost of some redundant computations. Specifically, for each PD, we project the PS of every individual point in the diagram onto a spline basis (without applying a weighting function). This design allows us to later change the weighting function $\omega$ and, by linearity, reuse the existing projections without repeating the spline projection step. While this is less efficient than projecting the entire PS directly, the latter approach would require reprojecting whenever $\omega$ changes. Striking the right balance between these two strategies, depending on the size of the parameter space, could lead to further efficiency improvements.

\section{Proofs of the Results}

\begin{prop}
Set $\lambda(p) := \frac{y-x}{2\parallel (1,p) \parallel_2}$. The following are stable weightings:
\[
\widetilde{\omega}(p)=\lambda(p)^\alpha, 
\qquad
\omega_K(p)=\frac{2}{\pi}\arctan \left(\frac{\lambda(p)^\alpha}{K^\alpha} \right), 
\]
 for any $K>0$ and $\alpha\geq1$. They are also effective weightings for $\alpha=1$. 
 \begin{proof}
 
The functions $\Gamma_\omega$ have the following forms:
\[
\Gamma_{\widetilde{\omega}}(x,y)= \frac{(y-x)^\alpha}{2^\alpha \parallel (1,x,y) \parallel_2^{\alpha}}  (1,x,y);
\]
\[
\Gamma_{\omega_k}(x,y)= \frac{2}{\pi}\arctan \left(\frac{(y-x)^\alpha}{2^\alpha K^\alpha \parallel (1,x,y) \parallel_2^{\alpha}} \right)  (1,x,y) .
\]
 
     Lipschitzianity is obtained because the components of the functions $\Gamma_{\widetilde{\omega}}$ and $\Gamma_{\omega_K}$ are differentiable and have bounded partial derivatives on $\R^2_{x<y}$. 

     To check the norm condition for stability, we write down the expressions of $\parallel \Gamma_\omega \parallel_2$:
\[
\parallel\Gamma_{\widetilde{\omega}}(x,y)\parallel_2= \frac{(y-x)^\alpha}{2^\alpha \parallel (1,x,y) \parallel_2^{\alpha-1}};
\]
     \[
\parallel\Gamma_{\omega_k}(x,y)\parallel_2= \frac{2}{\pi}\arctan \left(\frac{(y-x)^\alpha}{2^\alpha K^\alpha \parallel (1,x,y) \parallel_2^{\alpha}} \right) \parallel (1,x,y) \parallel_2.
\]

     At this point, we observe that:

     \[
     \frac{(y-x)^{\alpha-1}}{2^{\alpha-1}\parallel (1,x,y) \parallel_2^{\alpha-1}} \in [0,1);
     \]
     and that:
     \[
     \arctan \left(\frac{(y-x)^\alpha}{2^\alpha K^\alpha \parallel (1,x,y) \parallel_2^{\alpha}} \right)  \parallel (1,x,y) \parallel_2 \leq \frac{(y-x)^\alpha}{2^\alpha K^\alpha \parallel (1,x,y) \parallel_2^{\alpha-1}}.
     \]
     The first observation is enough to prove stability for $\widetilde{\omega}$, while the second and the first observations, combined, prove it for $\omega_K$.

     Now we prove that both weightings are effective for $\alpha=1$, exploiting \Cref{eq:effective}. 
     Note that the functions have become:  
\[
\parallel\Gamma_{\widetilde{\omega}}(x,y) \parallel_2 = \frac{(y-x)}{2}= \parallel p-\Delta \parallel_\infty;
\]
\[
\parallel\Gamma_{\omega_K}(x,y) \parallel_2 = \frac{2}{\pi}\arctan \left(\frac{(y-x)}{2 K \parallel (1,x,y) \parallel_2} \right) \parallel (1,x,y) \parallel_2.
\]     
To see that $\Gamma_{\widetilde{\omega}}$ is effective, it suffices to observe that, plugging the expression of $\parallel\Gamma_{\widetilde{\omega}} \parallel_2$ in \Cref{eq:effective}, we directly obtain the thesis.

Now we deal with $\Gamma_{\omega_K}$. Set $\mu_{D_n} = \sum_{p\in D_n} a_{n,p} \delta_p$. 

We rewrite \Cref{eq:effective} as:
\[
\lim_{r\rightarrow \infty} \sup_n \sum_{p\in D_n, \parallel p\parallel_2>r} \parallel  \Gamma_{\omega_K}(p) \parallel_2 \leq \lim_{r\rightarrow \infty} \sup_n \sum_{p\in D_n, \parallel p\parallel_2>r} a_{n,p} \parallel  \Gamma_{\omega_K}(p) \parallel_2\rightarrow 0.    
\]
Thus, for every $\varepsilon>0$, there is $R>0$ such that, for every $r>R$ the following holds:

\begin{equation}\label{eq:gamma}
 \sup_n \sum_{p\in D_n, \parallel p\parallel_2>r}\parallel  \Gamma_{\omega_K}(p) \parallel_2  < \varepsilon.        
\end{equation}

Hence, for every $n$, we have:
\begin{equation}\label{eq:arctan}
\sum_{p=(x,y)\in D_n, \parallel p\parallel_2>r} \arctan \left(\frac{(y-x)}{2 K \parallel (1,x,y) \parallel_2} \right)  <\frac{\pi \varepsilon}{2 r}.    
\end{equation}

In particular, \Cref{eq:arctan} implies $\frac{(y-x)}{2 K \parallel (1,x,y) \parallel_2}\rightarrow 0$ for $r\rightarrow \infty$. Equivalently, for every $C>0$, there is $r_C$ such that $\frac{(y-x)}{2 K \parallel (1,x,y) \parallel_2} <C$.

The key observation now, is that, due to the concavity of $z\mapsto\arctan(z)$, which is easy to see due to the strict monotonicity of its derivative $\frac{1}{1+z^2}$, we have:
\[
\frac{\arctan(\varepsilon)}{\varepsilon}z\leq \arctan(z) \leq z
\]
for every $z\in [0,\varepsilon]$, and every fixed $\varepsilon\geq 0$. In fact, $z\mapsto\frac{\arctan(\varepsilon)}{\varepsilon}z$ is the straight line joining $(0,0)$ and $(\varepsilon, \arctan(\varepsilon)).$
Thus, for every $C>0$ and for every $r_C$ such that $\frac{(y-x)}{2 K \parallel (1,x,y) \parallel_2} <C$, we have for every $n$:
\[
C' \sum_{p\in D_n, \parallel p\parallel_2>r_C} \frac{a_{n,p}(y-x)}{2}\leq \sum_{p\in D_n, \parallel p\parallel_2>r_C} \frac{2 a_{n,p}}{\pi}\arctan \left(\frac{(y-x)}{2 K \parallel (1,x,y) \parallel_2} \right) \parallel (1,x,y) \parallel_2,
\]
with $C'=\frac{2 \arctan(C)}{\pi C K}$.

In particular, we can find $C'>0$ such that:
\[
C'\lim_{r\rightarrow \infty} \sup_n \pers(\mu_{D_n}) \leq \lim_{r\rightarrow \infty} \sup_n \sum_{p\in D_n, \parallel p\parallel_2>r} a_{n,p} \parallel  \Gamma_{\omega_K}(p) \parallel_2\rightarrow 0.
\]
And the thesis follows.
 \end{proof}
\end{prop}

\begin{teo}
    Let $\mu_D,\mu_D'$ be PDs and let $\omega:\R^2\rightarrow \R$  be a stable weighting.
    We have:
    \[
    d_H(Z_{\mu^{\omega}_D},Z_{\mu^{\omega}_{D'}}) \leq \max\{C,C'\} W_1(\mu_D,\mu_{D'}),
    \]
    with $C,C'>0$ being the stability constants of $\omega$. 
    \begin{proof}
        Consider $\mu_D=\sum_{p\in D }a_p \delta_p,\mu_{D'}=\sum_{q\in D' }b_q \delta_q$ and a partial matching $\gamma$ between them. Without loss of generality, suppose $C=C'$.
        
A generic point in $Z_{\mu^{\omega}_D}$ has the form:
\[
P = \sum_{p\in D} a_p  s_p  \Gamma_\omega(p)\in Z_{\mu^\omega_D},
\]
with $s_p\in [0,1]$. We start by considering $p\in D_\gamma$ and $P\in Z_{\mu^\omega_D}$ with the following form: 
\begin{equation}\label{eq:P}
P = \gamma_p  \Gamma_\omega(p).   
\end{equation}

Note that, by definition, $\gamma_p\in\N$ and $\gamma_p\geq 1$.

Consider the point:
\[
Q = \gamma_p  \Gamma_\omega(\gamma(p)).
\]
Since $\gamma_p\leq b_{\gamma(p)}$, there is $s\in [0,1]$ such that $s b_{\gamma(p)} = \gamma_p$. Thus,
$Q\in Z_{\mu^{\omega}_{D'}}$.

We have:

\[
\parallel P- Q \parallel_2 \leq \gamma_p  \parallel \Gamma_\omega(p)-\Gamma_\omega(\gamma(p))\parallel_2 \leq \gamma_p  C  \parallel p-\gamma(p)\parallel_2^\alpha.
\]

For $P$ in the form of \Cref{eq:P}, we define $\Phi(P):=Q$.

Consider now a generic $P\in Z_{\mu^\omega_D}$:
\[
P = \sum_{p\in D} a_p  s_p  \Gamma_\omega(p) = \sum_{p\in D_\gamma} \gamma_p  s_p  \Gamma_\omega(p)+\sum_{p\in D_\gamma} (a_p-\gamma_p)  s_p  \Gamma_\omega(p)+\sum_{p\in D-D_\gamma} a_p  s_p  \Gamma_\omega(p),
\]
We build $Q$ as follows:
\[
Q = \sum_{p\in D_\gamma} s_p  \Phi(\gamma_p  \Gamma_\omega(p)).
\]
We have:

\begin{equation}\label{eq:P-Q}
    \begin{aligned}
\parallel P- Q \parallel_2 \leq  &\sum_{p\in D_\gamma}  s_p\gamma_p  \parallel \Gamma_\omega(p)-\Gamma_\omega(\gamma(p))\parallel_2 +\sum_{p\in D_\gamma} s_p (a_p-\gamma_p)   \parallel\Gamma_\omega(p)\parallel_2+ \\
&\sum_{p\in D-D_\gamma} s_p  a_p   \parallel\Gamma_\omega(p)\parallel_2.
    \end{aligned}
\end{equation}

Plugging into \Cref{eq:P-Q} the following facts:
\begin{enumerate}
    \item $ \parallel\Gamma_\omega(p)\parallel_2 \leq C \parallel p-\Delta \parallel_\infty$;
    \item $\parallel  \parallel_2 \leq \sqrt{2} \parallel  \parallel_\infty$;
    \item $s_p\leq 1$ for every $p\in D$,
\end{enumerate}
we obtain:
\begin{equation*}
    \begin{aligned}
&\parallel P- Q \parallel_2 \leq \\
&C  \left(\sqrt{2}  \sum_{p\in D_\gamma}\gamma_p \parallel p-\gamma(p)\parallel_\infty  + \sum_{p\in D_\gamma} (a_p-\gamma_p)  \parallel p-\Delta \parallel_\infty + \sum_{p\in D-D_\gamma} a_p   \parallel p-\Delta \parallel_\infty \right).  
    \end{aligned}
\end{equation*}

Since we can do this construction for any partial matching $\gamma$, for every $P \in Z_{\mu^\omega_D}$, we found $Q$ such that:
\[
\parallel P- Q \parallel_2 \leq C  W_1(\mu_D,\mu_{D'}).
\]
Reversing the role of $\mu_D$ and $\mu_{D'}$ we obtain the thesis.
    \end{proof}
\end{teo}

\begin{teo}
    Let $\{\mu_{D_n}\}_{n\in \N}$ be a sequence of PDs such that $d_H(Z_{\mu^{\omega}_{D_n}},Z_{\mu^{\omega}_{D}}) \rightarrow 0$, with $\omega:\R^2\rightarrow \R$  being an effective weighting. Then $W_1(\mu_{D_n},\mu_D)\rightarrow 0$.
    \begin{proof}
    
         By \Cref{prop:wass_conv}, we know that $d_H(Z_{\mu^{\omega}_{D_n}},Z_{\mu^{\omega}_{D}}) \rightarrow 0$ implies $\mu^{\omega}_{D_n}\xrightarrow{w}\mu^{\omega}_{D}$ and that $\{\mu^{\omega}_{D_n}\}$ is uniformly integrable. We immediately have $\mu^{\omega}_{D_n}\xrightarrow{v}\mu^{\omega}_{D}$.
        We only need to check that  $\pers(\mu_{D_n})\rightarrow \pers(\mu_D)$.

        Set $\mu_{D_n} = \sum_{p\in D_n} a_{n,p} \delta_p$. 

        By uniform integrability we have:
        \begin{equation*}
            \begin{aligned}
        &\lim_{r\rightarrow \infty}\sup_{n\in \N} \int_{B^c_{r}} \omega(p)\parallel p\parallel_2 d\mu_{D_n}(p) = \lim_{r\rightarrow \infty}\sup_{n\in \N} \int_{B^c_{r}} \parallel p\parallel_2 d\mu^\omega_{D_n}(p) = 0.
            \end{aligned}
        \end{equation*}

        Since $\omega$ is an effective weight, this implies:
        \[
        \lim_{r\rightarrow \infty}\sup_{n\in \N} \textstyle\pers_{B^c_{r}}(\mu_{D_n})\rightarrow 0,
        \]
        
        For any $r\geq 0$ we can write:
        \begin{equation*}
            \begin{aligned}
        &\pers(\mu_{D_n}) = \int_{\R^2_{x<y}} (y-x) d\mu_{D_n}((x,y)) =\int_{B^c_{r}} (y-x) d\mu_{D_n}((x,y))+ 
        \int_{B_{r}} (y-x) d\mu_{D_n}((x,y)).
            \end{aligned}
        \end{equation*}

        Similarly, we can write:
        \begin{equation*}
            \begin{aligned}
        \pers(\mu_{D}) =\int_{\R^2_{x<y}} (y-x) d\mu_D((x,y)) =
        \int_{B^c_{r}} (y-x) d\mu_D((x,y)) +
        \int_{B_{r}} (y-x) d\mu_D((x,y)).
            \end{aligned}
        \end{equation*} 
        If $r$ is big enough, being $D$ finite, we have $\text{supp}(D)\subset B_r$, and so
        $\int_{B_{r}^c} (y-x) d\mu_D((x,y))=0$ and $\int_{B_{r}} (y-x) d\mu_D((x,y))=\pers(\mu_D)$. 
        
        Fix some $r$ big enough so that the above holds.
        Since $B_{r}$ is compact we can write a positive test function $g:\R^2_{x<y}\rightarrow [0,1]$ such that:
        \begin{itemize}
            \item $g$ is continuous; 
            \item $g\equiv 1$ on $B_{r}$;
            \item $\text{supp}(g)$ is compact.
        \end{itemize}

        For such a $g$, we obtain:
        \[
        0\leq \int_{B_{r}} (y-x) d\mu_{D_n}((x,y)) \leq \int_{\R^2_{x<y}} g(x,y)(y-x) d\mu_{D_n}((x,y)).
        \]
        Moreover, using vague convergence, we get:
        \[
         \int_{\R^2_{x<y}} g(x,y)(y-x) d\mu_{D_n}((x,y))\xrightarrow{n} \pers(\mu_D).
        \]
        
        But:
        \begin{equation*}
            \begin{aligned}
         &\int_{\R^2_{x<y}} g(x,y)(y-x) d\mu_{D_n}((x,y)) = \\
         &\int_{B_r} g(x,y)(y-x) d\mu_{D_n}((x,y))+\int_{B_r^c} g(x,y)(y-x) d\mu_{D_n}((x,y))= \\
         &\int_{B_r} (y-x) d\mu_{D_n}((x,y))+\int_{B_r^c} g(x,y)(y-x) d\mu_{D_n}((x,y)).
            \end{aligned}
        \end{equation*} 
        Thus:
        \begin{equation*}
            \begin{aligned}
        & 0\leq \int_{\R^2_{x<y}} g(x,y)(y-x) d\mu_{D_n}((x,y)) -  \int_{B_r} (y-x) d\mu_{D_n}((x,y))  =   \\ 
        &\int_{B_r^c} g(x,y)(y-x) d\mu_{D_n}((x,y)) \leq 
        \int_{B_r^c} (y-x) d\mu_{D_n}((x,y)).
            \end{aligned}
        \end{equation*} 

    Putting the pieces together, for every $\varepsilon>0$ there exist $r_\varepsilon$ and $N_\varepsilon$ such that, for every $n\geq N_\varepsilon$:
    \[
    \mid \pers(\mu_{D_n}) - \int_{B_{r_\varepsilon}} (y-x) d\mu_{D_n}((x,y)) \mid \leq \varepsilon,
    \]
    \[
    \mid \pers(\mu_{D}) - \int_{\R^2_{x<y}} g(x,y) (y-x) d\mu_{D_n}((x,y)) \mid \leq  \varepsilon,
    \]
    \[
    0\leq \int_{\R^2_{x<y}} g(x,y) (y-x) d\mu_{D_n}((x,y)) - \int_{B_{r_\varepsilon}} (y-x) d\mu_{D_n}((x,y))  \leq \varepsilon,
    \]
    entailing $\mid \pers(\mu_{D_n}) - \pers(\mu_{D}) \mid \leq 3 \varepsilon$, concluding the proof.

    \end{proof}
\end{teo}

\end{document}